\pdfoutput=1

\documentclass[11pt]{article}

\usepackage[preprint]{acl}

\usepackage{times}
\usepackage{latexsym}

\usepackage[T1]{fontenc}

\usepackage[utf8]{inputenc}

\usepackage{microtype}

\usepackage{inconsolata}

\usepackage{graphicx}

\usepackage{amsmath}
\usepackage{amssymb}
\usepackage{mathtools}
\usepackage{amsthm}

\usepackage{adjustbox}

\title{Needle in the Haystack for Memory Based Large Language Models}

\author{
  \textbf{Elliot Nelson$^1$},
  \textbf{Georgios Kollias$^2$},
  \textbf{Payel Das$^3$},
  \textbf{Subhajit Chaudhury$^4$},
  \textbf{Soham Dan$^5$}
\\
  IBM Research 
}

\newcommand{\zkey}{\tilde{\mathbf{z}}}
\newcommand{\zb}{\mathbf{z}}
\newcommand{\wb}{\mathbf{w}}
\newcommand{\Zb}{\mathbf{Z}}
\newcommand{\Wb}{\mathbf{W}}
\newcommand{\Mb}{\mathbf{M}}
\newcommand{\Cb}{\mathbf{C}}
\newcommand{\Mkey}{\tilde{\mathbf{M}}}

\newcommand{\new}[1]{\textcolor{black}{#1}}
\newcommand{\cyan}[1]{\textcolor{black}{#1}}

\begin{document}

\maketitle

\footnotetext[1]{\href{mailto:enelson@ibm.com}{enelson@ibm.com}}
\footnotetext[2]{\href{mailto:gkollias@us.ibm.com}{gkollias@us.ibm.com}}
\footnotetext[3]{\href{mailto:daspa@us.ibm.com}{daspa@us.ibm.com}}
\footnotetext[4]{\href{subhajit@ibm.com}{subhajit@ibm.com}}
\footnotetext[5]{\href{mailto:soham.dan@ibm.com}{soham.dan@ibm.com}}

\begin{abstract}
Current large language models (LLMs) often perform poorly on simple fact retrieval tasks. Here we investigate if coupling a dynamically adaptable external memory to a LLM can alleviate this problem. For this purpose, we test Larimar, a recently proposed language model architecture which uses an external associative memory, on long-context recall tasks including passkey and needle-in-the-haystack tests. We demonstrate that the external memory of Larimar, which allows fast write and read of an episode of text samples, can be used at test time to handle contexts much longer than those seen during training. We further show that the latent readouts from the memory (to which  long contexts are written) control the decoder towards generating correct outputs, with the memory stored off of the GPU. Compared to existing transformer-based LLM architectures for long-context recall tasks that use larger parameter counts or modified attention mechanisms, a relatively smaller size Larimar is able to maintain strong performance without any task-specific training or training on longer contexts.
\end{abstract}

\section{Introduction}

One of the most important abilities of large language models (LLMs) is retrieving information from the text input that is included in the prompt, which enables generating contextually grounded responses. The length of the context an LLM can process during inference is therefore an important factor controlling the generation quality. Vanilla transformer models suffer from quadratic memory and computation complexity with respect to the length of the input sequence due to the self-attention mechanism. Further, the global and local information get mixed and \textit{blurred} while processing long context. Recent works have suggested different attention mechanisms in transformer-based LLM architectures to address such issues with long-context modeling \cite{munkhdalai2024leave, hwang2024transformerfam, beltagy2020longformer, dai2019transformerxl, RMT2022}. For example, transformers with Sliding Window Attention have been proposed that show $O(L\times W)$ complexity for input length $L$ and window size $W$ \cite{beltagy2020longformer}. \textit{Memory} of the past segments has been stored in a cache to be used as a context for the next segment \cite{dai2019transformerxl} or models have been trained to learn explicit \textit{global memory} tokens aka. soft prompts \cite{burtsev2021memory, RMT2022, hwang2024transformerfam}. Recently, Infini-transformer \cite{munkhdalai2024leave} has been proposed to use \textit{memory} of all past segments, as opposed to considering only the memory of the last segment in processing the current segment and discarding all other past segments. Those works often require task-specific training on long-context instances and lack generalization.

As an alternative, here we investigate test-time long-context adaptation of \textsc{larimar}, a recently proposed \cite{das2024larimar} LLM decoder model augmented with dynamically updatable memory mechanisms, and test its in-context recall performance on long-context modeling tasks.
\new{The external memory is structured similarly to the Kanerva Machine \cite{wu2018kanerva}, but updates the memory by finding the least-squares solution to a linear system, following \cite{pham2021generative}, instead of updating a multivariate Gaussian posterior distribution.}
While the model's training used relatively short contexts, we show that it can generalize to much longer contexts when only a small part of the context is task-relevant. This is because the external memory can be enlarged at test time to store and retrieve information from arbitrarily long contexts. 
Even if the context length vastly exceeds the training distribution, our model will generalize to the task as long as the memory readout (a single encoding) is in distribution as an input to the decoder.


In this paper, we make the following contributions:
\begin{itemize}
    \item We introduce a method for writing long contexts to an external associative memory, with read/write memory operations that 
    use a prefix or subset of the written text to ease retrieval when reading.
    \item We show that this method is able to perform long-context retrieval tasks at context length 100K-1M tokens, without any task-specific training, with the training context length limited to only 384 tokens, and with a relatively small 1.3B parameter model. 
    \item By performing all memory operations on the CPU, we demonstrate the feasibility of scaling to longer contexts without increasing the GPU memory space footprint.
\end{itemize}

\section{Background}

In this section, we review our model architecture and describe memory operations used for long-context recall tasks.

\textit{Notation.} We use lower-case characters to denote individual key or value vectors: $\zb\in\mathbb{R}^{C}$, $\wb\in\mathbb{R}^{K}$, with latent embedding dimension $C$ and memory size $K$, 
and upper-case for matrices of $N\geq1$ keys or values: $\Zb\in\mathbb{R}^{N \times C}$, $\Wb\in\mathbb{R}^{N \times K}$.


\subsection{Larimar Architecture Overview}
\label{sec:larimar}

The language model (LM) employed here is Larimar \cite{das2024larimar}, which uses an encoder-decoder architecture \cite{li2020optimus} trained together 
with a linear associative memory module \cite{pham2021generative}.
The LM encoder and decoder are trained to reconstruct an episode of input text samples as follows:  The text is encoded in the latent space and written to the memory, the memory is queried, and the readout from the memory conditions the decoder (see \cite{das2024larimar} for details). \new{(Note that the decoder receives as input text a query about the context, but only accesses the full context -- which is stored in the external memory -- via the memory readout channel.)} 
The loss objective used is a combination of a cross-entropy reconstruction loss and an auto-encoder loss. 
During inference, the context is divided into $N$ segments, each of which are encoded and then written to memory. In our experiments, we use the following memory operations, differing slightly from \cite{das2024larimar}.\footnote{We also set $\sigma_{\Wb}=\sigma_\xi=0$ as defined in \cite{das2024larimar}, resulting in deterministic read and write operations.}

\textit{Writing.}
To write a segment of text to memory, we first compute its encoding $\zb$, along with a writing key vector $\wb$ defined below. 
Given arrays of new encodings $\Zb$ and corresponding key vectors $\Wb$, the memory matrix $\Mb\in\mathbb{R}^{K\times C}$ is obtained as the solution\footnote{$\Wb^{\dagger}$ indicates the pseudoinverse of the non-square matrix $\Wb$.}
\begin{equation} \label{eq:M_write}
    \Mb = \Wb^{\dagger}\Zb
\end{equation}
to the least-squares problem of minimizing the readout error $||\Zb-\Wb\Mb||_2^2$ 
\cite{kohonen1973,bau2020rewriting}. 

\textit{Reading.} We assume the context ends in a final query (partial sentence), which is treated differently from the preceding context. Instead of writing to memory, the query is used to compute a reading key $\wb$, with which an encoding
\begin{equation}
    \label{eq:z_read}
    \zb_{\rm read} = \wb\Mb
\end{equation}
is read out from memory and passed as input to the decoder.

\textit{Reading and Writing Keys.} The reading and writing keys used with a given encoding $\zb$ are computed as a function $\wb = f(\zkey|\Mkey)$ of an encoding $\zkey$ (which may differ from the encoding $\zb$ written to memory), conditional on a fixed ``key memory'' $\Mkey\in\mathbb{R}^{K\times C}$ (distinct from $\Mb$) which is used exclusively to compute key vectors $\wb$.
The encoding $\zkey$ can be obtained using a fixed-length \textit{prefix} of the text to be written (when writing) or the query text (when reading), or alternatively can be the full text encoding $\zb$. Using a prefix can lead to reading and writing key vectors which are more similar.\footnote{For instance, if the written text is ``The pass key is 732'' and the query text is ``The pass key is'', using the same three- or four-word prefix of each will result in equivalent reading and writing keys.}
The function $f$ is defined as follows.
Given the encoding $\zkey$, we select the nearest neighbor row in the key memory matrix,
\begin{equation}
     k^\star(\zkey) := \text{argmin}_k ||\zkey - \Mkey_{k,:}||_2.
     \label{eq:nearest_neighbor_slot}
\end{equation}
The corresponding one-hot key vector is
\begin{equation}
    \label{eq:w_one_hot}
    w_k = \mathbf{1}(k=k^\star(\zkey)),
\end{equation}
where $\mathbf{1}(x)=1$ when $x$ is True, and $0$ otherwise. 
This ensures that the rows of $\Mb$ in Eq. \eqref{eq:M_write} are simply the encodings $\Zb$, and that the memory readout vector in Eq. \eqref{eq:z_read} is simply the $k^\star(\zkey)$'th row of $\Mb$, that is, $\zb_{\rm read} = \Mb_{k^\star(\zkey),:}$.
Lastly, we set the rows of the key memory $\Mkey$ to the prefix encodings $\zkey$ of each in-context sentence. This ensures that, in the limit where the query prefix encoding $\zkey=\zkey_{\rm query}$ is very close to a particular prefix encoding used when writing to memory (e.g. $\zkey=\zkey_{\rm needle}$), the nearest neighbor row, $k^\star(\zkey)$ is in fact the row of $\Mb$ where the corresponding sentence (e.g. $\zb=\zb_{\rm needle}$) was written.

In general, the time complexity of writing to memory is set by the $O(K^3)$ cost of the pseudoinverse $\Wb^\dagger$. However, when key vectors are one-hot, Eq. \eqref{eq:w_one_hot}, and furthermore when the nearest neighbor locations $k^\star(\zkey)$ are unique for each segment encoding $\zkey$ (as in the case where these encodings populate the rows of $\Mkey$) and the memory size $K$ is set to the $O(N)$ number of unique segments, then $\Wb$ is a permutation of the identity matrix. In this case, and with the $O(N)$ computations of key vectors running in parallel, the overall runtime of computing $\Mb$ (as well as reading from it) is $O(N)$.

\section{Experiments}

\cyan{We conducted two experiments to evaluate long-context recall. These experiments collectively involved $O$(10) GPU hours with an A100 40GB GPU.}


\hspace{-0.38cm}\textbf{Passkey test.}

We test Larimar on the passkey test as defined in \cite{mohtashami2023landmark}.\footnote{The exact context is as follows: ``There is an important info hidden inside a lot of irrelevant text. Find it and memorize them. I
will quiz you about the important information there. The grass is green. The sky is blue. The sun
is yellow. Here we go. There and back again. (\textit{repeat x times}) The pass key is 9054. Remember
it. 9054 is the pass key. The grass is green. The sky is blue. The sun is yellow. Here we go.
There and ack again. (\textit{repeat y times}) What is the pass key? The pass key is''} 
The context is divided into sentences, which are each written to memory, with the exception of the final prompt ``The pass key is,'' which is fed into the decoder along with the memory readout.

In Table~\ref{table:passkey_9054} we report the the average retrieval accuracy compared to the Zero-shot accuracy of Infini-Transformer \cite{munkhdalai2024leave}. 
Importantly, because the context has only a small number of distinct sentences, regardless of the context length, only a small number of memory slots are used. All repeats of a given sentence are written to the same memory slot $k^\star(\zb)$ in Eq. \eqref{eq:nearest_neighbor_slot}.
Consequently, the memory readout and model generation are independent of the context length; while we only report up to 1M tokens in Table \ref{table:passkey_9054}, the same results will hold for arbitrarily long contexts. (This will not hold for contexts where the number of unique text segments grows with context length.)
While Table~\ref{table:passkey_9054} assumes a single random passkey number, we also evaluated (at 1.2M tokens) the average recall rate over 100 random numbers with $n$ digits, with results shown in Table \ref{table:passkey}.

Note that our method is invariant to the order in which sentences are written to memory, resulting in equivalent performance for any position of the passkey (or needle sentence, below) within the context.

\hspace{-0.38cm}\textbf{Needle-in-the-Haystack.}

We follow \cite{needlehaystack_kamradt}, using the ``haystack'' dataset of Paul Graham essays, for which the total context length is $\approx137$K tokens.
We test completion of needle sentences of the form ``The magic number is X'' with final prompt ``The magic number is'' following the context, as well as the ``San Francisco'' needle\footnote{The needle text is ``The best thing to do in San Francisco is eat a sandwich and sit in Dolores Park on a sunny day.'' After writing the context to memory we use the prompt ```The best thing to do in San Francisco is ''.} \cite{needlehaystack_kamradt}.

For Larimar, when computing key vectors, we used the encoding $\zkey$ of the four-word prefix of each sentence.\footnote{Or the full sentence, if it was shorter.}
We set the memory size $K$ to the total number of sentences $N$, and write each encoded sentence to a unique memory slot using the key vector method of section \ref{sec:larimar}. This incurs a memory storage cost that scales linearly with the context length, similar to \cite{kuratov2024search}. However, by keeping the external memory and read/write operations on the CPU (with encoding and memory-conditioned decoding on the GPU), we avoid an increased GPU memory cost and are able to handle much longer context lengths. 

\cyan{We compare to Mistral 7B v0.2 and Phi-3-Mini-128K-Instruct (3.8B parameters) as baseline methods for long-context recall. For the Mistral model, we use 1200-sentence ($\approx24$K tokens) subsets 
of the full haystack dataset to evaluate the model at slightly less than its 32K token context limit. For the Phi-3 model, we use a 5000-sentence subset ($\approx100K$ tokens). 
We average over needle positions distributed uniformly throughout the context. 
}

Our results are reported in Table \ref{table:needle}, and show Larimar's ability to maintain strong recall at over 100K context tokens, while baseline models struggle with shorter contexts. 
For Larimar, note that the drop in recall from 3 to 4 digits reflects the increased difficulty of reconstructing the needle from the original encoding, rather than an increase in difficulty in locating the needle encoding in memory.
\cyan{The benefit of using a shorter, fixed-length prefix to compute the writing key is more significant for longer or more complex needle sentences (4-digit numbers, SF needle) in which case the query encoding may differ more from the full, untruncated needle encoding. (When the full needle encoding is too different from the query encoding, the reading key finds a different ``haystack'' sentence which is closer in the latent space.)}

\begin{table}[!ht]
\centering
\begin{center}
\begin{adjustbox}{width=0.46\textwidth}    
    \begin{tabular}{ l  l  l}
    \hline
    & 128K & 1M \\ \hline
    Larimar  & 100/100/100 & 100/100/100 \\
    Infini-Transformer (Linear) & 11/14/100 & 8/6/98 \\
    Infini-Transformer (Linear + $\delta$) & 6/9/99 & 7/6/97 \\ \hline
    \end{tabular}
\end{adjustbox}    
\end{center}
\caption{Percentage of successful recall for Larimar \cyan{1.3B} on the passkey test as defined in Appendix B of \cite{munkhdalai2024leave}}
\label{table:passkey_9054}
\end{table}

\begin{table}[!ht]
\centering
\begin{center}
\begin{adjustbox}{width=0.14\textwidth}    
    \begin{tabular}{ l  l  l  l}
    \hline
    \textbf{Passkey} & 1.2M \\ \hline
    3 digits & 100 \\
    4 digits & 86 \\
    5 digits & 87 \\
    6 digits & 79 \\
    7 digits & 59 \\
    8 digits & 50 \\ \hline
    \end{tabular}
\end{adjustbox}    
\end{center}
\caption{Percentage of successful recall for Larimar 1.3B on the passkey test with context length up to 1,200,057, averaged over $100$ random passkey numbers in each case.}
\label{table:passkey}
\end{table}

\begin{table}[!ht]
\centering
\begin{center}
\begin{adjustbox}{width=0.45\textwidth}    
    \begin{tabular}{ l | l | l  l l}
    \hline
     & context & 3 digits & 4 digits & SF  \\ \hline
    Larimar & 137K & 0.95 & 0.64 & 1.0 \\
    Larimar (no prefix) & 137K & 0.88 & 0.14 & 0.0  \\
    Mistral 7B v0.2 & 24K & 0.66 & 0.62 & 0.80 \\
    Phi-3-mini-128K & 100K & 0.27 & 0.26 & 0.37 \\
    \end{tabular}
\end{adjustbox}    
\end{center}
\caption{Percentage of successful recall for Larimar 1.3B and baseline models on the needle-in-the-haystack test. The ``context'' column indicates the context length used, with the remaining columns showing results for different ``needle'' sentences. 
For 3 or 4 digit ``magic number'' needles, we average over $50$ random numbers, checking the presence of the number within the model response.
For the ``San Francisco'' needle, we evaluate the rougeL recall (length of the longest common subsequence of words, divided by the length of the target completion).
For Larimar, we either truncate each sentence to a fixed-length prefix when computing writing keys, or do not (``no prefix'').
}
\label{table:needle}
\end{table}

\section{Discussion}


Recent approaches to long-context retrieval have shown good performance after fine-tuning smaller models on needle-in-the-haystack tasks. 
RMT \cite{RMT2022} and RMT-Retrieval \cite{kuratov2024search} have shown near-perfect recall out to $1$M tokens, with task-specific training of GPT2 (137M parameters).
Infini-attention \cite{munkhdalai2024leave} with fine-tuning on the passkey test obtains $100\%$ 
recall with context lengths up to $1$M tokens. 
Without fine-tuning, however, recall drops to $O(10\%)$ at context length $128$K (Table~\ref{table:passkey_9054}). 

On the other hand, larger models 
have shown strong performance on needle-in-the-haystack tests without task-specific fine-tuning, but incur additional training and inference costs due to their larger size.

In comparison to these approaches, we aim for strong recall performance with a more compact model (1.3B parameters), without resorting to task-specific training. 
Our model was trained on a generic text dataset using a subset of Wikipedia data, and can be adapted to novel context data during inference, with the reduced latency and inference costs of a smaller model \cite{das2024larimar}.


In our experiments, we allow the memory space to grow in proportion with the context length.
While this increases space complexity, we emphasize that it does not increase the storage space needed on the GPU, since all memory operations can be performed on one or more CPUs. (A single long-context query requires (i) the encoded context to be moved to the CPU for writing to memory, and (ii) the memory readout to be moved back to the GPU for decoding, with the decoder input sequence length being $\lesssim O(100)$ tokens regardless of the context length.)
Overall, we emphasize that the external memory size can be adjusted as needed depending on the task and context, with the memory-conditioned decoder training allowing the model to adapt to variable-sized contexts with unseen data. 
\cyan{
Deepening memory hierarchy in hardware by adding disk storage to CPU RAM can further expand available space and flexibility in offloading limited GPU memory \cite{sheng2023flexgen}. 
}


In the future, we aim to explore more general methods for computing reading and writing keys conditional on the full context, such that memory space can be dynamically allocated to context data that is more task-relevant and/or more surprising to the model, allowing for more predictable parts of the context to be stored in memory with correspondingly fewer bytes of information.

\section{Limitations}

An algorithmic limitation of our approach is that, after dividing the context into segments (e.g. sentences), each segment is written to memory in isolation, losing the information in cross-segment correlations and the sequence order of segments. Our method is thus most relevant for tasks where the relevant information is within individual segments. It could also be incorporated into more general architectures that extract context information in long-range correlations before writing to an external memory.


\bibliography{refs}

\begin{thebibliography}{16}
\providecommand{\natexlab}[1]{#1}

\bibitem[{Bau et~al.(2020)Bau, Liu, Wang, Zhu, and Torralba}]{bau2020rewriting}
David Bau, Steven Liu, Tongzhou Wang, Jun-Yan Zhu, and Antonio Torralba. 2020.
\newblock \href {https://arxiv.org/abs/2007.15646} {Rewriting a deep generative model}.
\newblock \emph{Preprint}, arXiv:2007.15646.

\bibitem[{Beltagy et~al.(2020)Beltagy, Peters, and Cohan}]{beltagy2020longformer}
Iz~Beltagy, Matthew~E. Peters, and Arman Cohan. 2020.
\newblock \href {https://arxiv.org/abs/2004.05150} {Longformer: The long-document transformer}.
\newblock \emph{Preprint}, arXiv:2004.05150.

\bibitem[{Bulatov et~al.(2022)Bulatov, Kuratov, and Burtsev}]{RMT2022}
Aydar Bulatov, Yury Kuratov, and Mikhail Burtsev. 2022.
\newblock Recurrent memory transformer.
\newblock In \emph{Advances in Neural Information Processing Systems}, volume~35, pages 11079--11091.

\bibitem[{Burtsev et~al.(2021)Burtsev, Kuratov, Peganov, and Sapunov}]{burtsev2021memory}
Mikhail~S. Burtsev, Yuri Kuratov, Anton Peganov, and Grigory~V. Sapunov. 2021.
\newblock \href {https://arxiv.org/abs/2006.11527} {Memory transformer}.
\newblock \emph{Preprint}, arXiv:2006.11527.

\bibitem[{Dai et~al.(2019)Dai, Yang, Yang, Carbonell, Le, and Salakhutdinov}]{dai2019transformerxl}
Zihang Dai, Zhilin Yang, Yiming Yang, Jaime Carbonell, Quoc~V. Le, and Ruslan Salakhutdinov. 2019.
\newblock \href {https://arxiv.org/abs/1901.02860} {Transformer-xl: Attentive language models beyond a fixed-length context}.
\newblock \emph{Preprint}, arXiv:1901.02860.

\bibitem[{Das et~al.(2024)Das, Chaudhury, Nelson, Melnyk, Swaminathan, Dai, Lozano, Kollias, Chenthamarakshan, Jiří, Navrátil, Dan, and Chen}]{das2024larimar}
Payel Das, Subhajit Chaudhury, Elliot Nelson, Igor Melnyk, Sarath Swaminathan, Sihui Dai, Aurélie Lozano, Georgios Kollias, Vijil Chenthamarakshan, Jiří, Navrátil, Soham Dan, and Pin-Yu Chen. 2024.
\newblock \href {https://arxiv.org/abs/2403.11901} {Larimar: Large language models with episodic memory control}.
\newblock \emph{Preprint}, arXiv:2403.11901.

\bibitem[{Hwang et~al.(2024)Hwang, Wang, Huo, Sim, and Mengibar}]{hwang2024transformerfam}
Dongseong Hwang, Weiran Wang, Zhuoyuan Huo, Khe~Chai Sim, and Pedro~Moreno Mengibar. 2024.
\newblock \href {https://arxiv.org/abs/2404.09173} {Transformerfam: Feedback attention is working memory}.
\newblock \emph{Preprint}, arXiv:2404.09173.

\bibitem[{Kamradt(2023)}]{needlehaystack_kamradt}
Gregory Kamradt. 2023.
\newblock \href {https://github.com/gkamradt/LLMTest_NeedleInAHaystack/tree/main} {{Needle In A Haystack} - pressure testing {LLM}s}.
\newblock \emph{Github}.

\bibitem[{Kohonen and Ruohonen(1973)}]{kohonen1973}
Teuvo Kohonen and Matti Ruohonen. 1973.
\newblock \href {https://doi.org/10.1109/TC.1973.5009138} {Representation of associated data by matrix operators}.
\newblock \emph{IEEE Trans. Comput.}, 22(7):701–702.

\bibitem[{Kuratov et~al.(2024)Kuratov, Bulatov, Anokhin, Sorokin, Sorokin, and Burtsev}]{kuratov2024search}
Yuri Kuratov, Aydar Bulatov, Petr Anokhin, Dmitry Sorokin, Artyom Sorokin, and Mikhail Burtsev. 2024.
\newblock \href {https://arxiv.org/abs/2402.10790} {In search of needles in a 11m haystack: Recurrent memory finds what llms miss}.
\newblock \emph{Preprint}, arXiv:2402.10790.

\bibitem[{Li et~al.(2020)Li, Gao, Li, Peng, Li, Zhang, and Gao}]{li2020optimus}
Chunyuan Li, Xiang Gao, Yuan Li, Baolin Peng, Xiujun Li, Yizhe Zhang, and Jianfeng Gao. 2020.
\newblock Optimus: Organizing sentences via pre-trained modeling of a latent space.
\newblock In \emph{Proceedings of the 2020 Conference on Empirical Methods in Natural Language Processing (EMNLP)}, pages 4678--4699.

\bibitem[{Mohtashami and Jaggi(2023)}]{mohtashami2023landmark}
Amirkeivan Mohtashami and Martin Jaggi. 2023.
\newblock \href {https://arxiv.org/abs/2305.16300} {Landmark attention: Random-access infinite context length for transformers}.
\newblock \emph{Preprint}, arXiv:2305.16300.

\bibitem[{Munkhdalai et~al.(2024)Munkhdalai, Faruqui, and Gopal}]{munkhdalai2024leave}
Tsendsuren Munkhdalai, Manaal Faruqui, and Siddharth Gopal. 2024.
\newblock \href {https://arxiv.org/abs/2404.07143} {Leave no context behind: Efficient infinite context transformers with infini-attention}.
\newblock \emph{Preprint}, arXiv:2404.07143.

\bibitem[{Pham et~al.(2021)Pham, Le, Ngo, Tran, Ho, and Venkatesh}]{pham2021generative}
Kha Pham, Hung Le, Man Ngo, Truyen Tran, Bao Ho, and Svetha Venkatesh. 2021.
\newblock Generative pseudo-inverse memory.
\newblock In \emph{International Conference on Learning Representations}.

\bibitem[{Sheng et~al.(2023)Sheng, Zheng, Yuan, Li, Ryabinin, Chen, Liang, R{\'e}, Stoica, and Zhang}]{sheng2023flexgen}
Ying Sheng, Lianmin Zheng, Binhang Yuan, Zhuohan Li, Max Ryabinin, Beidi Chen, Percy Liang, Christopher R{\'e}, Ion Stoica, and Ce~Zhang. 2023.
\newblock Flexgen: high-throughput generative inference of large language models with a single gpu.
\newblock In \emph{Proceedings of the 40th International Conference on Machine Learning}, pages 31094--31116.

\bibitem[{Wu et~al.(2018)Wu, Wayne, Graves, and Lillicrap}]{wu2018kanerva}
Yan Wu, Greg Wayne, Alex Graves, and Timothy Lillicrap. 2018.
\newblock \href {https://arxiv.org/abs/1804.01756} {The kanerva machine: A generative distributed memory}.
\newblock \emph{Preprint}, arXiv:1804.01756.

\end{thebibliography}

\appendix

\section{Potential Risks}
\label{sec:risks}

Our paper describes an approach to long-context recall for language models using an external memory. Language models with increased memory and recall capabilities introduce potential risks via misuse, and should only be deployed with appropriate guardrails. 
Our experiments only involved publicly available data without sensitive information, and we only applied our method on a relatively small 1.3B parameter model, with lower capability levels and potential for misuse compared to larger language models.

\end{document}